\documentclass[journal]{IEEEtran}
\usepackage{graphicx}
\usepackage{float}
\begin{document}
\pagenumbering{gobble}
\title{A Scalable and Robust Framework for Intelligent Real-time Video Surveillance}

\author{Shreenath~Dutt, Ankita~Kalra\\%
		\textit{Department~of~Electronics~Engineering}\\%
		\textit{Indian Institute Of Technology~(BHU), Varanasi, India}\\
		\textit{shreenath.dsharma.ece11@iitbhu.ac.in}\\
		\textit{ankita.kalra.ece11@iitbhu.ac.in}
		}

\maketitle

\begin{abstract}
In this paper, we present an intelligent, reliable and storage-efficient video surveillance system using Apache Storm and OpenCV. As a Storm topology, we have added multiple information extraction modules that only write important content to the disk. Our topology is extensible, capable of adding novel algorithms as per the use case without affecting the existing ones, since all the processing is independent of each other. This framework is also highly scalable and fault tolerant, which makes it a best option for organisations that need to monitor a large network of surveillance cameras.\end{abstract}


\section{Introduction}
The rise in surveillance camera deployments is making a big financial impact on public as well as private sector'€™s operating expenditures. The growing number of cameras, increasing video quality, and the government compliance of retaining videos for longer periods of time, have caused a surge in storage space requirements. The real issue is not the quality or the number of videos, but that the cameras are recording 24 hours a day regardless of whether something is happening in the field of view. We plan to tackle this problem by Intelligent Video surveillance, which not only reduces the storage requirements, but also makes the review easier, by storing important information separately.\\
The outline of our paper is as follows: In following subsections, we discuss about distributed stream processing and the limitations of using \textit{Apache Hadoop} for the same. We then introduce \textit{Apache Storm} as the low latency framework for distributed computation. In Section II, we explain the architecture of Apache Storm and the data model used by StormCV for Video Processing. In Section III, we present our workflow for Intelligent Video Surveillance along with the graphical representation of our topology. In Section IV we show the results of our algorithm on one of the datasets. Section V provides a conclusion and explores the future scope.

\subsection{Distributed Stream Processing}
Stream processing is an old idea, but it is currently being rediscovered in industry due to increasing data volumes and diverse data sources. To cope up with the amount of data, Apache Hadoop is being used for distributed processing of large datasets across clusters of computers. It is designed to scale up from single server to thousands of machines leveraging their own resources. However, MapReduce is inherently designed for high throughput batch processing of big data, which makes it difficult to be used for low latency applications.

\subsection{Apache Storm}
Apache Storm \cite{apache-storm} is a distributed computation framework, which uses \textit{spouts} and \textit{bolts} to define information sources and processing operations to allow for distributed processing of streaming data. It is designed as a Directed Acyclic graph with Spouts and bolts as its nodes. The Edges are called as Streams, which transfer data from one node to the other. This arrangement of Spouts and Bolts is called a Topology.

\section{Apache Storm Framework}
A Storm cluster is superficially similar to a Hadoop cluster. The \textit{Topologies} used in Storm are analogous to Hadoop \textit{Jobs}. However, while a Map Reduce Job eventually finishes, Storm Topologies can run indefinitely until killed.
There are two kinds of nodes on a Storm cluster: the master node and the worker node. The master node runs a daemon called \textit{Nimbus}, which is responsible for distributing code around the cluster, assigning tasks to machines, and monitoring for failures. Each worker node runs a daemon called the \textit{Supervisor}, which listens for work assigned by Nimbus and starts and stops worker processes as necessary. All coordination between Nimbus and the Supervisors is done through a Zookeeper cluster.

\begin{figure}
\centering
\includegraphics[width=2.5in,clip,keepaspectratio]{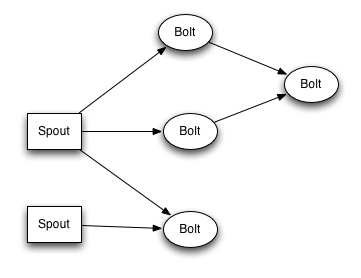}
\caption{Storm Topology}
\label{fig_sim}
\end{figure}

The core abstraction of Storm is a \textit{Stream}, which is an unbounded sequence of tuples. Spouts serve as source of streams, which can emit multiple streams at a time. Similarly, a bolt can consume multiple streams at once, do some processing and possibly emit new streams to be processed further. When a spout or bolt emits a tuple to a stream, it sends the tuple to every bolt that subscribed to that stream.
Each node in a Storm Topology has a capability to execute in parallel. In case of any failed task, Storm reassigns it automatically, ensuring that there is no data loss. This feature makes Storm a reliable stream processing framework.

\subsection{StormCV}
StormCV is an extension of Apache Storm designed to support the development of distributed computer vision processing pipelines \cite{stormcv-github}. It enables the use of Apache Storm by providing computer vision specific operations and data model. Its data model uses Frames, Features and Descriptor objects that are serializable to Storm tuples. StormCV topologies are created by defining and linking a number of  Spouts/Bolts and specifying which Fetchers/Operations they should execute. Fetchers are responsible for reading imaging data and are executed within Spouts. Operations take one or more objects from the model and produce zero or more output objects. They can, for example, receive a Frame and generate multiple Feature objects.

\section{Intelligent Video Surveillance Model}
We use StormCV for developing our distributed pipeline for processing the video feed. Following are the components of this topology -
\subsection*{Input}
\subsection{StreamFrameFetcher (Spout)}
 We receive the video feed frame by frame using \textit{\texttt{StreamFrameFetcher}} interface of StormCV, which is responsible for reading frames from a source at specific intervals, and emitting them as Frame objects. We emulated the real-time video stream by streaming a webcam on localhost using Real Time Streaming Protocol (RTSP), which was read directly by the Spout executing the Fetcher.
 \subsection*{Information Extraction}
 \subsection{Background Subtraction Bolt}
 This bolt consumes frames from the StreamFrameFetcher [\textit{Sec.III(A)}] spout and does further processing. Since our main motive is to avoid storing background's static information, we subtract background from each incoming frame. For this, we implemented a Moving Average based background subtraction algorithm, using OpenCV's \textit{\texttt{accumulateWeighted}} \cite{opencv-weightedAccumulated} function to calculate the moving average. Since it is a moving average, we don't need to provide a ground truth in the beginning for the accumulator to learn. As the time passes and we feed more and more incoming frames, it dynamically learns the background distribution. This also makes it adaptable to varying background conditions. After subtracting an incoming frame from the background, we get a grayscale difference image which we then threshold using a primitive thresholding function. Then we calculate the blob sizes to avoid false positives like motion of leaves and slight illumination changes. Only those frames which have a blob area more than 10\% of frame area, were labelled as having a foreground. These frames are then passed to next bolt which processes them and writes them to a folder, which will eventually be used to create a video feed.

 \subsection{Face Detection Bolt}
 Depending upon the use case, we can deploy different bolts and have them process information at several stages. We propose to store the faces recognised from the video feed separately in chronologically arranged folders. After consuming frames from the Background Subtraction Bolt~[\textit{Sec.III(B)}], we used OpenCV's pre-trained Haar Cascades along with the \textit{detectMultiScale} function to detect faces. The cascade that we have currently used is trained to detect frontal faces. The \textit{detectMultiScale} function in OpenCV uses Viola Jones detection algorithm with Adaboost Classifier.\cite{opencv-vj}.  After processing, the bolt adds face descriptor information in the frame and pass es that to the next node of the stream in the topology.

\subsection{Person Detection Bolt}
 Since frontal view of face might not always be visible for detection,  we have added an additional bolt for detecting persons in the Video Stream. Similar to the Face Detection Bolt [\textit{Sec. III(C)}], this bolt consumes frames from the Background Subtraction Bolt~[\textit{Sec.III(B)}, and uses OpenCV's pre-trained HOG Cascades with the \textit{detectMultiscale} function to detect persons. The cascade used in this bolt detects person from its front, side and back-view. Hence, there is no obligation for presence of a clear facial image in the video. After descriptors are generated, it adds them to the frames and generates a stream that is processed by the labeller Bolt in the next node of the topology.
 
 \subsection*{Labeling and Output Generation}
 \subsection{Labeller Bolt}
 The Labeller Bolt consumes streams from both the Detection Bolts~[\textit{Sec. III(C,D)}] and labels the bounding boxes as per the descriptor received in respective frames' features. After labelling the frames, it submits them to the next bolt in the topology, which writes these frames to the disk.

 \subsection{Export to File Bolt}
The Export Bolt writes the received frames from Labeller Bolt and Background Subtraction Bolt~[\textit{Sec. III(E,B)}] to the disk according to the streams a frame is part of. For the purpose of the paper, we stored the three information from the Video stream in three separate folders -\\
\begin{itemize}
	\item One for the frames that contained any foreground information, as transmitted by the Background Subtraction Bolt. This Folder~(\textit{EligibleFrames}) will be used to create the Video chunks to be written to the disk.
	\item Other Two for storing the labeled frames containing detected Face and Persons from the video feed.
\end{itemize}

\subsection{Export to Video Bolt}
This Bolt runs independently from all the other streams, and reads directly from the StreamFrameFetcher~[\textit{Sec.III(A)}]. It does not do any processing on the frames \textit{per se}, but it keeps polling the \textit{EligibleFrames} folder until there are \textit{N} number of frames to be written before creating a video. This number(\textit{N}) can be configured depending upon the  duration of individual video chunks to be created.
\begin{center}
\begin{math}
	N = Frame~Rate(fps) * Desired~Video~length(s)
\end{math}	
\end{center}
After reading files from the folder, we have used Xuggler's \textit{IMediaWriter Interface}~\cite{xuggler} to write to the video with a constant frame rate.
\section*{Graphical Overview of the Topology}
\begin{figure}[H]
\centering
\includegraphics[width=3.5in,clip,keepaspectratio]{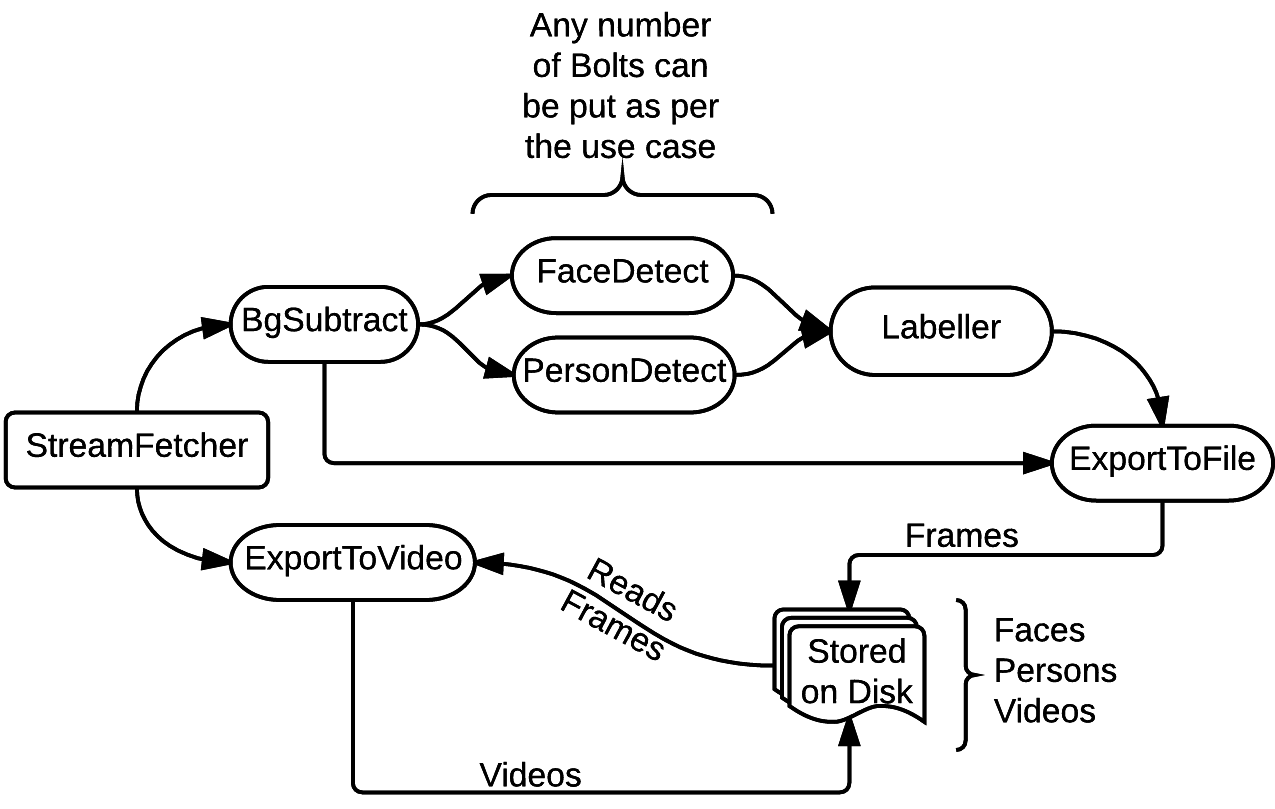}
\caption{Proposed Topology}
\label{fig_sim}
\end{figure}
The Storm topology is fault tolerant and guarantees message processing, i.e. if there are faults during execution of the computation, Storm reassigns tasks as necessary. This is established via Storm's \textit{reliability} API \cite{storm-rel-api}.
Additionally, Storm topologies are easily scalable. To scale a topology, we just need to add new machines and increase the parallelism settings of the topology. The usage of Zookeeper for cluster coordination allows it to scale to much larger cluster sizes. This robustness makes it a best option to be used for organisations that need to manage a large network of Surveillance Cameras.

\section{Results}
Since video surveillance feeds from different sources have different content, hence the distribution of empty frames varies with the use case. Therefore, reduction in the final size of the video is dependent on the video source. For instance, a surveillance feed of a traffic junction will rarely have an empty frame. On the other hand, surveillance feeds of ATM cubicles, office and hotel lobbies, home security cameras do not witness a lot of commotion throughout the day. For these places, we would see a considerable reduction in video size written to disk as compared to video feeds from former sources.
Also, since we are removing any frame that does not contain a foreground object, the resulting video will always be lesser in size than the original one.\\
For comparison, we submitted the image sequences from various camera feeds from the \textit{HDA+ Person Dataset} \cite{hda} to our topology and noted the size of the output. One of the Image sequences in the same dataset, \textit{(Camera18)} consisted of 9882 frames, taking up 83.8MB of disk space. After processing the frames, we filtered out 1418 frames that contained information, eventually taking up 11.9MB disk space. This resulted in a 85.65\% decrease in disk usage. Further results are explained in the following table -\\

{\tiny
\def\arraystretch{1.5}
\begin{center}
    \begin{tabular}{|l|l|r|r|r|r|r|}
      \hline
        &&\multicolumn{2}{c|}{\centering \textit{Pre Processing}} & \multicolumn{2}{c|}{\centering \textit{Post Processing}}& \\
      \cline{3-6}
      \multicolumn{1}{|c|}{\textit{Video}} & \multicolumn{1}{|c|}{\textit{Resolution}} & \multicolumn{1}{c|}{\textit{No. of frames}} & \multicolumn{1}{c|}{\textit{Video Size}} & \multicolumn{1}{c|}{\textit{No. of frames}}& \multicolumn{1}{c|}{\textit{Video Size}} & \multicolumn{1}{|c|}{\textit{\textit{\% Reduction}}}\\
      \hline
      \textit{Camera02} & \textit{640~x~480} & 9818   & 47.1MB & 1248   & 5.9MB & 87.28\%\\
      \textit{Camera17} & \textit{640~x~480} & 9896   & 42.0MB & 1791   & 7.6MB & 81.90\%\\
      \textit{Camera18} & \textit{640~x~480} & 9882   & 83.8MB & 1418   & 11.9MB & 85.65\%\\
      \textit{Camera19} & \textit{640~x~480} & 9887   & 59.7MB & 2934   & 17.7MB & 70.32\%\\
      \textit{Camera40} & \textit{640~x~480} & 9860   & 37.0MB & 2790   & 10.4MB & 71.70\%\\
      \textit{Camera50} & \textit{1280~x~800} & 2226   & 38.5MB & 662    & 11.46MB & 70.26\%\\
      \hline
    \end{tabular}
  \end{center}
}

The \textit{\%Reduction}, however, depends a lot on video content. Since above videos were taken in office and lobby settings, we saw a good improvement on the storage size on application of our topology.
\section{Conclusion and Future Scope}
In this paper we have presented an intelligent video surveillance system using Apache Storm. It improves on the disk usage than many of the currently used surveillance systems, by only storing the relevant content to the disk. We also extracted the important information such as faces and the persons detected from the video and made it easily accessible for it to be reviewed later, without having to browse through the entire video.\\
Benefits of such a system are manyfold. This would help average households economically, since the camera retention time (time taken for the disk to get full) will increase, hence decreasing the frequency of changing storage disks.  
Using Apache Storm for the framework makes it reliable and easily scalable, we only have to add new feeds in the Stream Fetcher and the topology assigns them different \textit{StreamIds} for distinguishing frames emanating from different cameras.\\
As for the information extraction is considered, we now have an opportunity to implement any of the complicated information extraction algorithms in realtime by assigning different \textit{StreamIds} to the streams coming from different modules, so that all the streams are independent of each other and computational time of one of the streams does not affect the other.

\end{document}